\documentclass{article}

\usepackage{microtype}
\usepackage{graphicx}
\usepackage{subcaption}
\usepackage{booktabs} 
\usepackage{placeins}

\usepackage{hyperref}

\usepackage[preprint]{icml2026}

\usepackage{amsmath}
\usepackage{amssymb}
\usepackage{mathtools}
\usepackage{amsthm}

\usepackage[capitalize,noabbrev]{cleveref}

\theoremstyle{plain}

\theoremstyle{definition}

\theoremstyle{remark}

\usepackage[textsize=tiny]{todonotes}

\icmltitlerunning{Why the Maximum Second Derivative of Activations Matters for Adversarial Robustness}

\usepackage{tikz}
\usetikzlibrary{arrows.meta, positioning, shapes.misc}
\usepackage{mathtools}
\usepackage{amssymb}
\usepackage{amsmath}
\usepackage{bm,yhmath}
\usepackage{nicefrac} 
\usepackage{algpseudocode}
\usepackage{adjustbox,multirow,siunitx}
\usepackage{amsthm}
\usepackage{microtype}
\usepackage{graphicx}
\usepackage{subcaption}
\usepackage{booktabs} 
\usepackage{makecell}

\usepackage[capitalize,noabbrev]{cleveref}

\newcommand\yuaf{{\text{RCT-AF}}}

\makeatletter%
\newcolumntype{L}{D{.}{.}{2,2}}
\newcolumntype{B}[3]{>{\boldmath\DC@{#1}{#2}{#3}}c<{\DC@end}}

\begin{document}

\twocolumn[
  \icmltitle{Why the Maximum Second Derivative of Activations Matters for Adversarial Robustness}



  \icmlsetsymbol{equal}{*}

  \begin{icmlauthorlist}
  \icmlauthor{Yunrui Yu}{thu}
  \icmlauthor{Hang Su}{thu}
  \icmlauthor{Jun Zhu}{thu}
  \end{icmlauthorlist}

\icmlaffiliation{thu}{Tsinghua University, Beijing, China}

\icmlcorrespondingauthor{Yunrui Yu}{yuyunrui@mail.tsinghua.edu.cn}
\icmlcorrespondingauthor{Hang Su}{suhangss@mail.tsinghua.edu.cn}
\icmlcorrespondingauthor{Jun Zhu}{dcszj@mail.tsinghua.edu.cn}

  \icmlkeywords{Machine Learning, ICML}

  \vskip 0.3in
]



\printAffiliationsAndNotice{}  

\begin{abstract}
This work investigates the critical role of activation function curvature---quantified by the maximum second derivative $\max|\sigma''|$---in adversarial robustness. Using the Recursive Curvature-Tunable Activation Family (RCT-AF), which enables precise control over curvature through parameters $\alpha$ and $\beta$, we systematically analyze this relationship. Our study reveals a fundamental trade-off: insufficient curvature limits model expressivity, while excessive curvature amplifies the normalized Hessian diagonal norm of the loss, leading to sharper minima that hinder robust generalization. This results in a non-monotonic relationship where optimal adversarial robustness consistently occurs when $\max|\sigma''|$ falls within 4 to 10, a finding that holds across diverse network architectures, datasets, and adversarial training methods. We provide theoretical insights into how activation curvature affects the diagonal elements of the hessian matrix of the loss, and experimentally demonstrate that the normalized Hessian diagonal norm exhibits a U-shaped dependence on $\max|\sigma''|$, with its minimum within the optimal robustness range, thereby validating the proposed mechanism.
\end{abstract}    
\section{Introduction}

Deep neural networks have demonstrated remarkable success across a wide range of domains, yet their vulnerability to adversarial attacks remains a critical challenge \cite{szegedy2013intriguing, goodfellow2014explaining}. Adversarial training has emerged as one of the most effective defense mechanisms, where models are trained on adversarially perturbed examples to enhance robustness \cite{madry2017towards, zhang2019theoretically}. While substantial research has focused on developing more sophisticated training algorithms, the influence of architectural components—particularly activation functions—on adversarial robustness has received comparatively limited systematic investigation.

Activation functions play a foundational role in determining the expressive capacity and optimization dynamics of neural networks. The widespread adoption of the non-smooth ReLU \cite{nair2010rectified} has been complemented by the development of smooth, twice-differentiable alternatives such as GELU \cite{hendrycks2016gaussian} and Swish \cite{ramachandran2017searching}, which often yield improved performance in standard training tasks. A defining characteristic of these smooth activations is their non-vanishing second derivative, which typically attains a maximum magnitude $\max|\sigma''|$ around zero. While prior work suggests smooth activations can benefit adversarial training \cite{xie2020smooth}, the systematic role and optimal magnitude of $\max|\sigma''|$ in adversarial robustness remain unexplored.

In this work, we systematically investigate how the maximum second derivative of activation functions affects adversarial robustness. Using the \yuaf\ , which provides a principled framework for controlling $\max|\sigma''|$ through parameters $\alpha$ and $\beta$, we identify a fundamental trade-off via extensive empirical evaluation: insufficient $\max|\sigma''|$ (below $\sim$4) limits nonlinear expressivity, impairing model capacity; moderate $\max|\sigma''|$ (between 4 and 10) yields optimal adversarial robustness by balancing expressivity and loss landscape flatness; conversely, excessive $\max|\sigma''|$ (above $\sim$10) leads to reduced adversarial robustness, as observed consistently across experimental settings.


To explain this empirical observation, we derive the explicit mathematical relationship between the activation second derivative $\sigma''$ and the diagonal elements of the loss Hessian. Based on the Gauss-Newton decomposition, our derivation shows that $\sigma''$ contributes linearly to individual diagonal elements $[\nabla^2_\theta \hat{L}(\theta)]_{kk}$. Since these diagonal elements can have opposing signs, we examine the normalized Hessian diagonal $L_2$ norm $\|\operatorname{diag}(\nabla^2_\theta \hat{L})/p\|_2$, where $p$ is the total number of parameters. This norm avoids cancellation by summing squared values. Empirically, we find that the normalized Hessian diagonal norm exhibits a U-shaped dependence on $\max|\sigma''|$, attaining its minimum precisely within the optimal robustness range (4-10) and increasing outside it. This provides a direct mechanistic link: deviation from the optimal $\max|\sigma''|$ range increases the normalized Hessian diagonal norm, indicating a sharper loss landscape, which prior work associates with poorer robust generalization \cite{keskar2016large, foret2020sharpness}.

Extensive experiments on CIFAR-10 and CIFAR-100 with ResNet-18 and WideResNet-28-10 architectures, using diverse adversarial training methods including DAJAT \cite{addepalli2022efficient}, DKL \cite{cui2024decoupled}, and TRADES \cite{zhang2019theoretically}, validate these insights. We demonstrate that optimal adversarial robustness consistently occurs when $\max|\sigma''|$ falls within 4 to 10, a finding that holds across diverse network architectures, datasets, and adversarial training methods.

The primary contributions of this work are:
\begin{itemize}
    \item We systematically investigate the relationship between the maximum second derivative of activation functions and adversarial robustness, revealing a fundamental trade-off that yields an optimal range for $\max|\sigma''|$ (4 to 10).
    
    \item We derive the explicit mathematical relationship between the activation second derivative $\sigma''$ and the diagonal elements of the loss Hessian, and empirically demonstrate that the normalized Hessian diagonal norm exhibits a U-shaped dependence on $\max|\sigma''|$, with its minimum within the optimal robustness range, thereby explaining the observed non-monotonic robustness trend.
    
    \item Through extensive experiments with the \yuaf\ , we show that controlling $\max|\sigma''|$ within the optimal range significantly improves adversarial robustness across multiple datasets, network architectures, and adversarial training methods.
    
\end{itemize}

\section{Related Work}

\subsection{Adversarial Attacks and Robustness Evaluation}
The discovery of adversarial examples \cite{szegedy2013intriguing} initiated a sustained arms race between increasingly sophisticated attacks and defenses. Early gradient-based attacks, such as the Fast Gradient Sign Method (FGSM) \cite{goodfellow2014explaining} and its iterative variant Projected Gradient Descent (PGD) \cite{madry2017towards}, established the paradigm of perturbing inputs along the loss gradient. However, many proposed defenses were later found to rely on gradient masking or obfuscation, leading to inflated robustness estimates that were circumvented by stronger attacks \cite{athalye2018obfuscated}. This evaluation crisis underscored the need for rigorous, attack-agnostic benchmarks. The introduction of \textit{AutoAttack (AA)} \cite{croce2020reliable} addressed this by providing a reliable, parameter-free ensemble of diverse attack strategies, establishing it as a standard for robustness evaluation. 
In this work, we adopt AutoAttack to ensure a reliable assessment of robustness, enabling us to focus on intrinsic architectural factors.

\subsection{Adversarial Training Methods}
In response to powerful attacks, adversarial training has remained one of the most effective defense strategies. The foundational formulation by \cite{madry2017towards} minimizes the worst-case loss within a perturbation budget, though it often induces a trade-off between standard and robust accuracy. This work has inspired numerous algorithmic improvements: TRADES \cite{zhang2019theoretically} provides a theoretical decomposition of the loss; MART \cite{wang20misclass} emphasizes misclassification-aware optimization; DAJAT \cite{addepalli2022efficient} enhances robustness via efficient data augmentation; and DKL \cite{cui2024decoupled} employs a decoupled KL-divergence loss for stable training. While these methods primarily advance robust training at the *algorithmic* level, our work complements this line of research by investigating how intrinsic *architectural* components—specifically, the curvature properties of activation functions—fundamentally affect adversarial robustness.


\subsection{Activation Functions in Deep Learning}
Activation functions are fundamental architectural elements that determine neural networks' nonlinear expressivity. The Rectified Linear Unit (ReLU) \cite{nair2010rectified} revolutionized deep learning by mitigating vanishing gradients, but its non-differentiability at zero and complete suppression of negative signals can hinder optimization. Variants such as Leaky ReLU \cite{maas2013rectifier} and Parametric ReLU (PReLU) \cite{he2015delving} address the ``dead neuron'' issue by introducing a small, non-zero slope for negative inputs, yet remain non-smooth at zero. Truly smooth, twice-differentiable alternatives—including the Exponential Linear Unit (ELU) \cite{clevert2015fast}, Scaled ELU (SELU) \cite{klambauer2017self}, Gaussian Error Linear Unit (GELU) \cite{hendrycks2016gaussian}, Swish \cite{ramachandran2017searching}, and Mish \cite{misra2019mish}—have demonstrated superior performance in many standard training tasks, benefiting from improved gradient flow and higher-order differentiability.
In adversarial training, \cite{xie2020smooth} showed that smooth activations can improve robustness compared to ReLU, attributing this to better gradient propagation during adversarial example generation. However, these works have not systematically investigated the role of the activation function's second derivative $\sigma''$—particularly its maximum magnitude $\max|\sigma''|$—on adversarial robustness. The specific impact of activation curvature on loss landscape geometry and robust generalization remains unexplored.

\subsection{Loss Landscape Analysis and Hessian-based Metrics}
The geometry of the loss landscape has been closely linked to generalization performance. \cite{keskar2016large} empirically associated sharp minima with poorer generalization, while \cite{foret2020sharpness} proposed Sharpness-Aware Minimization (SAM) to explicitly optimize for flat minima. In adversarial robustness, \cite{wu2020adversarial} showed that flatter minima correlate with better robust generalization.

The Hessian matrix $\nabla^2_\theta L(\theta)$ serves as a fundamental tool for quantifying loss curvature. However, in non-convex landscapes, the trace $\operatorname{tr}(\nabla^2_\theta L)$ can be misleading due to cancellations between positive and negative eigenvalues. To avoid this issue and capture the overall curvature intensity, our work focuses on the \textit{diagonal elements} of the normalized Hessian and their $L_2$ norm $\|\operatorname{diag}(\nabla^2_\theta L)/p\|_2 = \sqrt{\frac{1}{p}\sum_k [\nabla^2_\theta L]_{kk}^2}$, where $p$ is the total number of parameters. This normalized diagonal norm avoids cancellation by summing squared values. 
\section{Methods}\label{sec:method}
\subsection{\yuaf\ : A Family of Activation Functions with Controllable Rectification and Curvature}

We leverage the Recursive Curvature-Tunable Activation Family (\yuaf\ ) family \cite{yu2025rcr}, a principled framework for generating activation functions with systematically controllable rectification strength and curvature properties. The \yuaf\  formulation is defined through a recursive operation $(\cdot)' \cdot x$ applied to a base function, generating increasingly nonlinear activation functions with adjustable parameters.

The base function ($\beta=0$) is defined as:
\begin{equation}
\begin{split}
&\text{\yuaf\ }(x; \alpha; \beta=0) \\
& = \frac{1}{\alpha}\ln\left(1+e^{\alpha x}\right),
\label{eq:beta0}
\end{split}
\end{equation}
which is a parameterized form of the softplus function \cite{glorot2011deep}. Its second derivative is:
\begin{equation}
\sigma''_{\beta=0}(x) = \frac{\alpha e^{-\alpha x}}{(1+e^{-\alpha x})^2}.
\label{eq:beta0_second}
\end{equation}
The maximum second derivative occurs at $x=0$ with value $\alpha/4$.

The first-order variant ($\beta=1$) is obtained through the operation $(\cdot)' \cdot x$:
\begin{equation}
\begin{split}
&\text{\yuaf\ }(x; \alpha; \beta=1)\\
&= \left(\text{\yuaf\ }(x; \alpha; \beta=0)\right)' \cdot x \\
&= \frac{x}{1+e^{-\alpha x}}, 
\label{eq:beta1} 
\end{split}
\end{equation}
which constitutes a parameterized version of the Swish activation function \cite{ramachandran2017searching}. Its second derivative is:
\begin{equation}
\sigma''_{\beta=1}(x) = \frac{\alpha e^{\alpha x}\left(e^{\alpha x}(2-\alpha x)+\alpha x+2\right)}{(e^{\alpha x}+1)^3}.
\label{eq:beta1_second}
\end{equation}
The maximum second derivative occurs at $x=0$ with value $\alpha/2$.

The second-order variant ($\beta=2$) is obtained through successive application:
\begin{equation}
\begin{split}
&\text{\yuaf\ }(x; \alpha; \beta=2) \\
&= \left(\left(\text{\yuaf\ }(x; \alpha; \beta=0)\right)' \cdot x\right)' \cdot x  \\
&= \frac{\left(e^{2\alpha x} + (\alpha x + 1)e^{\alpha x}\right)x}{e^{2\alpha x} + 2e^{\alpha x} + 1}, \label{eq:beta2}
\end{split}
\end{equation}
with second derivative:
\begin{equation}
\begin{split}
\sigma''_{\beta=2}(x) = \alpha e^{\alpha x}&\left((\alpha x-1)(\alpha x-4)e^{2\alpha x}\right.\\
&-4(\alpha^2 x^2-2)e^{\alpha x}\\
&\left.+(\alpha x+1)(\alpha x+4)\right)/(e^{\alpha x}+1)^4.
\label{eq:beta2_second}
\end{split}
\end{equation}
The maximum second derivative occurs at $x=0$ with value $\alpha$.

\begin{figure*}[h!]
    \centering%
    \begin{subfigure}{0.33\linewidth}
        \centering
        \includegraphics[width=\linewidth]{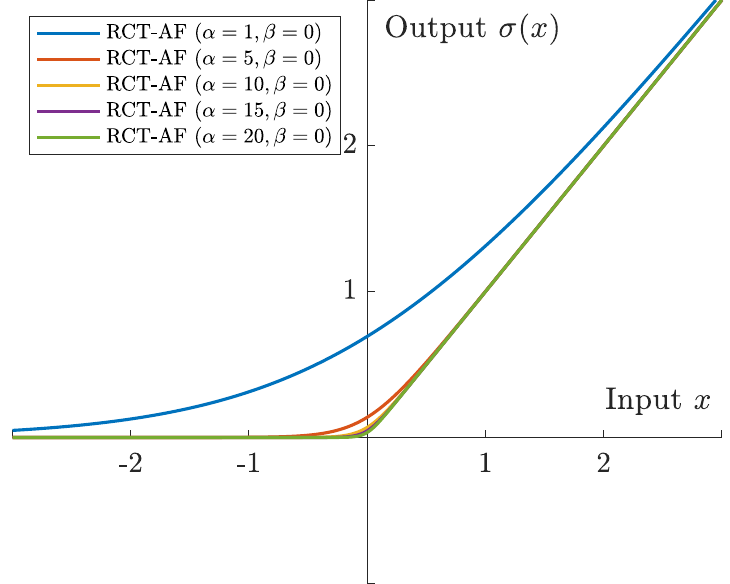}
        \caption{ $\beta=0$}
        \label{figa1}
    \end{subfigure}
    \begin{subfigure}{0.33\linewidth}
        \centering
        \includegraphics[width=\linewidth]{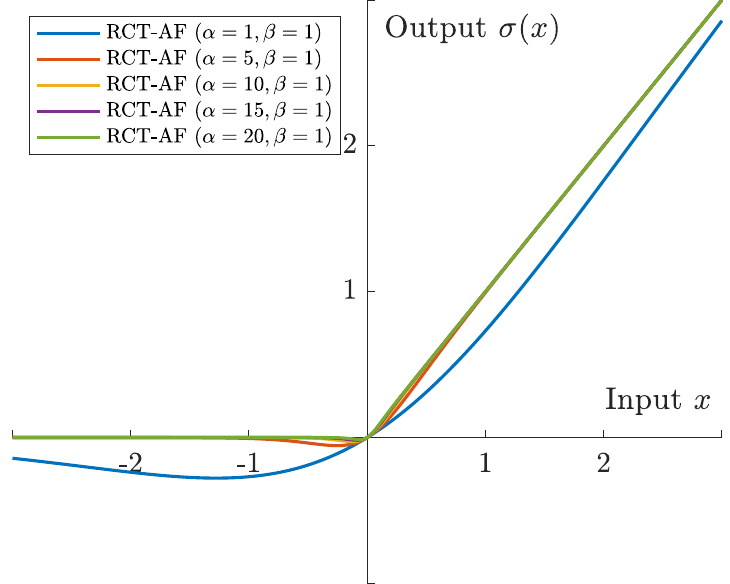}
        \caption{$\beta=1$}
        \label{figb1}
    \end{subfigure}
    \begin{subfigure}{0.33\linewidth}
        \centering
        \includegraphics[width=\linewidth]{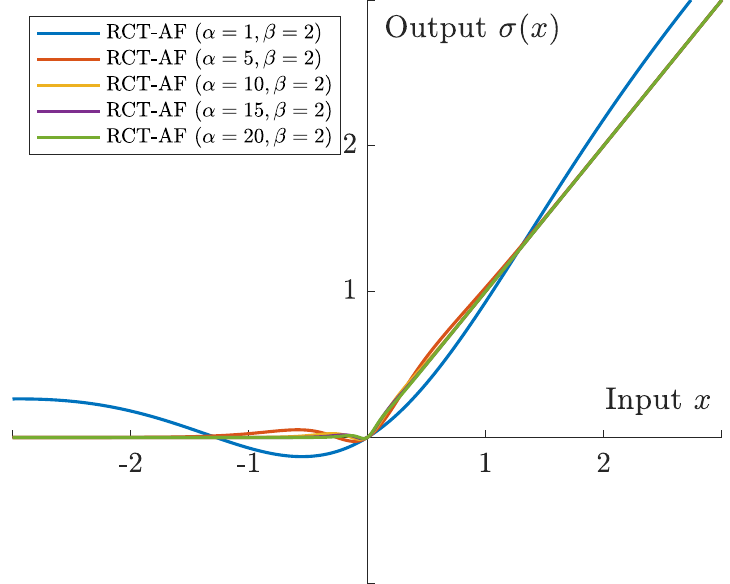}
        \caption{$\beta=2$}
        \label{figc1}
    \end{subfigure}
    \caption{%
Comparison of \yuaf\  activation functions with tunable rectification strength via $\alpha$ parameter ($\alpha = 1, 5, 10, 15, 20$). 
(a) $\beta = 0$: Baseline form.
(b) $\beta = 1$: First-order variant.
(c) $\beta = 2$: Second-order variant.
Across all configurations, increasing $\alpha$ systematically strengthens the asymmetric treatment of negative versus positive inputs, enabling precise control over the activation function's rectification characteristics.
}\label{fig:ablation_alpha_different_beta}
\end{figure*}

\begin{figure*}[h!]
    \centering%
    \begin{subfigure}{0.33\linewidth}
        \centering
        \includegraphics[width=\linewidth]{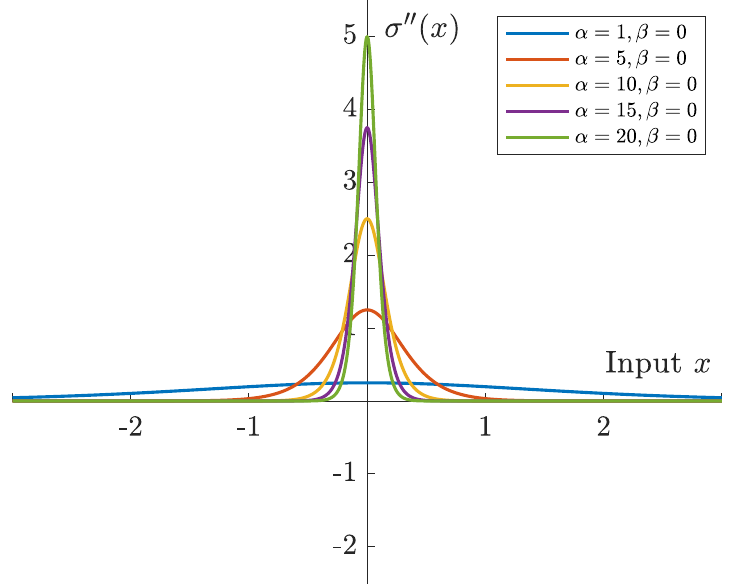}
        \caption{ $\beta=0$}
        \label{figa}
    \end{subfigure}
    \begin{subfigure}{0.33\linewidth}
        \centering
        \includegraphics[width=\linewidth]{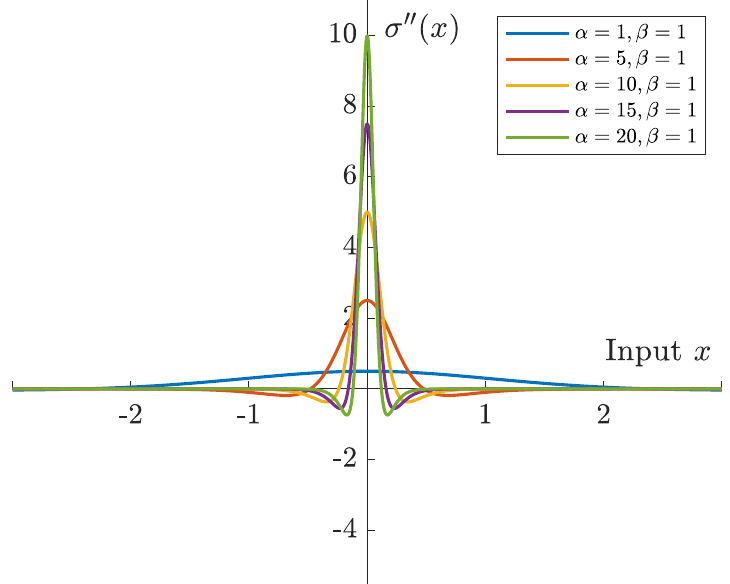}
        \caption{$\beta=1$}
        \label{figb}
    \end{subfigure}
    \begin{subfigure}{0.33\linewidth}
        \centering
        \includegraphics[width=\linewidth]{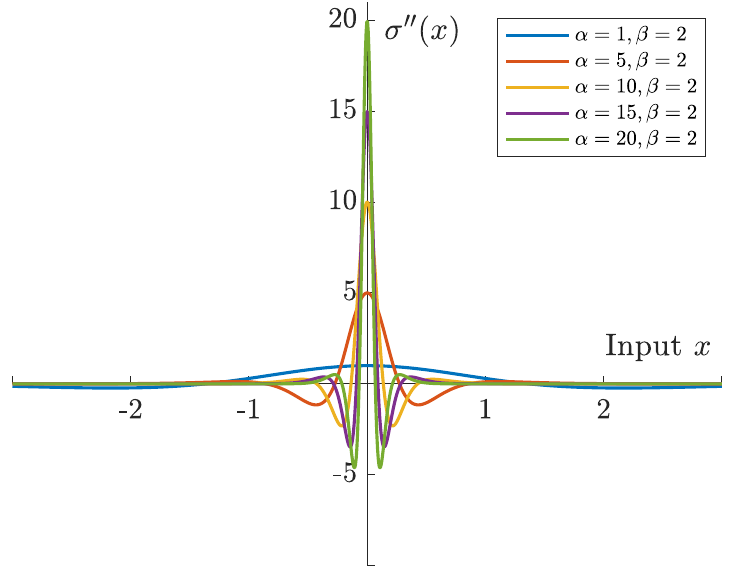}
        \caption{$\beta=2$}
        \label{figc}
    \end{subfigure}
    \caption{%
        Second derivative analysis of symmetric \yuaf\  activation functions for $\alpha \in \{1, 5, 10, 15, 20\}$. 
(a) $\beta = 0$: $\sigma''_{\beta=0}(x)$ yields a single extremum at $x=0$ with value $\alpha/4$.
(b) $\beta = 1$: $\sigma''_{\beta=1}(x)$ produces three critical points: a maximum at $x=0$ ($\alpha/2$) and two symmetric minima.
(c) $\beta = 2$: $\sigma''_{\beta=2}(x)$ exhibits five critical points: a central maximum at $x=0$ ($\alpha$) and four symmetric extremal points.
All functions are even and symmetric about the $y$-axis.}\label{fig:ablation_second_alpha_different_beta}
\end{figure*}

Figures~\ref{fig:ablation_alpha_different_beta} and \ref{fig:ablation_second_alpha_different_beta} visually demonstrate how $\alpha$ and $\beta$ control rectification strength and curvature. As shown in Figure~\ref{fig:ablation_alpha_different_beta}, increasing $\alpha$ systematically enhances the rectification strength across all $\beta$ values, leading to more pronounced asymmetric treatment of positive and negative inputs. Simultaneously, Figure~\ref{fig:ablation_second_alpha_different_beta} reveals that the maximum curvature, quantified by the peak second derivative at $x=0$, increases linearly with $\alpha$: $\alpha/4$ for $\beta=0$, $\alpha/2$ for $\beta=1$, and $\alpha$ for $\beta=2$. This provides a clean experimental setup where we can independently vary rectification strength (through $\alpha$) and functional form (through $\beta$) while precisely controlling the resulting curvature.

\subsection{Theoretical Analysis: Connecting Activation Second Derivatives to Hessian Diagonal Elements}

Our theoretical analysis reveals how the second derivative of activation functions influences the diagonal elements of the Hessian matrix of the loss. The derivation is presented for $L$-layer fully connected networks. Through the Gauss-Newton decomposition and careful chain rule application (see Appendix for complete derivation), we establish the explicit relationship between individual Hessian diagonal elements and activation second derivatives.

For any parameter $\theta_k$ associated with neuron $i$ in layer $l$, the corresponding Hessian diagonal element can be expressed as a function of activation second derivatives along all subsequent layers:
\begin{multline}
    [\nabla^2_\theta \hat{L}]_{kk} 
    = (\delta^{(l)}_i c_k)^2 + (f-y) c_k^2 \\
    \times \sum_{r=l}^{L-1} \sum_{\substack{\text{paths } P: \\ (l,i) \to (r,j)}} 
        \sigma''(z^{(r)}_j) \cdot \frac{\delta^{(r)}_j}{\sigma'(z^{(r)}_j)} \\
    \times \prod_{s=l}^{r-1} [\sigma'(z^{(s)}_{i_s})]^2 (W^{(s+1)}_{i_{s+1}, i_s})^2,
    \label{eq:diag_expanded_main}
\end{multline}
where $\delta^{(l)}_i = \partial f / \partial z^{(l)}_i$ is the backpropagated gradient, $c_k = \partial z^{(l)}_i / \partial \theta_k$ is a constant ($c_k = h^{(l-1)}_j$ for weights, $c_k = 1$ for biases), and the inner sum runs over all paths $P$ connecting neuron $(l,i)$ to neuron $(r,j)$.

Equation \eqref{eq:diag_expanded_main} demonstrates that the second derivative $\sigma''(z^{(r)}_j)$ directly contributes to the Hessian diagonal element $[\nabla^2_\theta \hat{L}(\theta)]_{kk}$. During adversarial training, the residual $(f-y)$ is typically significant, especially in early and middle training stages, making the influence of activation curvature substantial.

\begin{figure*}[h!]
\centering
\begin{subfigure}{0.49\textwidth}
\centering
\includegraphics[width=\linewidth]{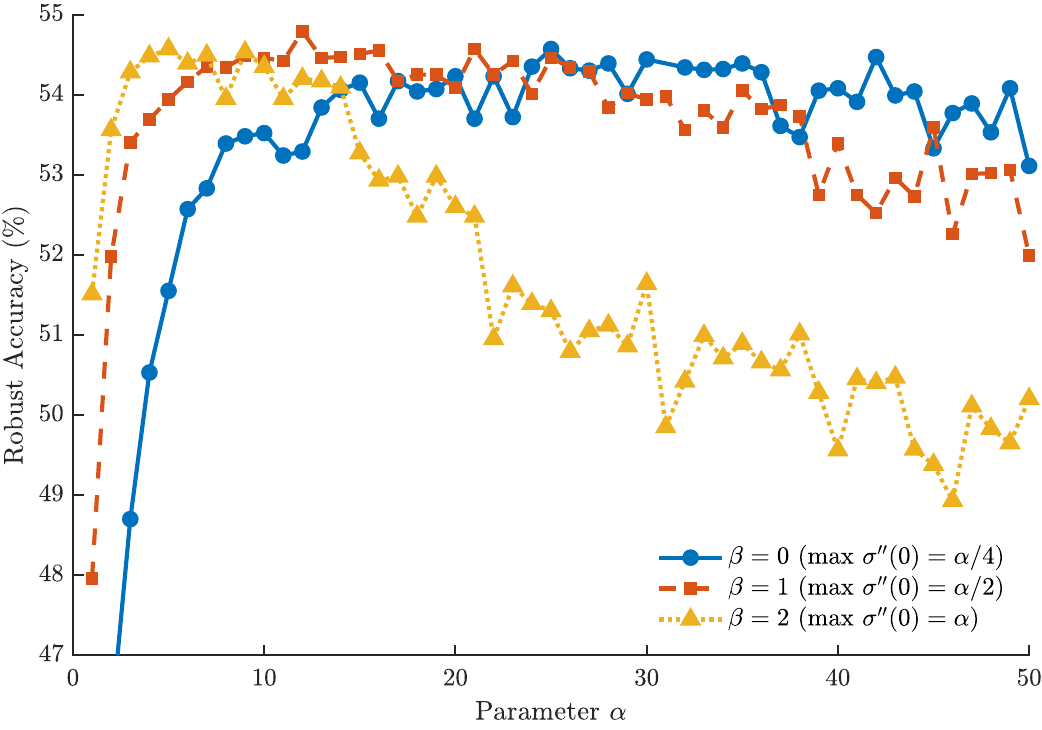}
\caption{Robustness vs. $\alpha$ parameter}
\label{fig:robustness_alpha}
\end{subfigure}
\hfill
\begin{subfigure}{0.49\textwidth}
\centering
\includegraphics[width=\linewidth]{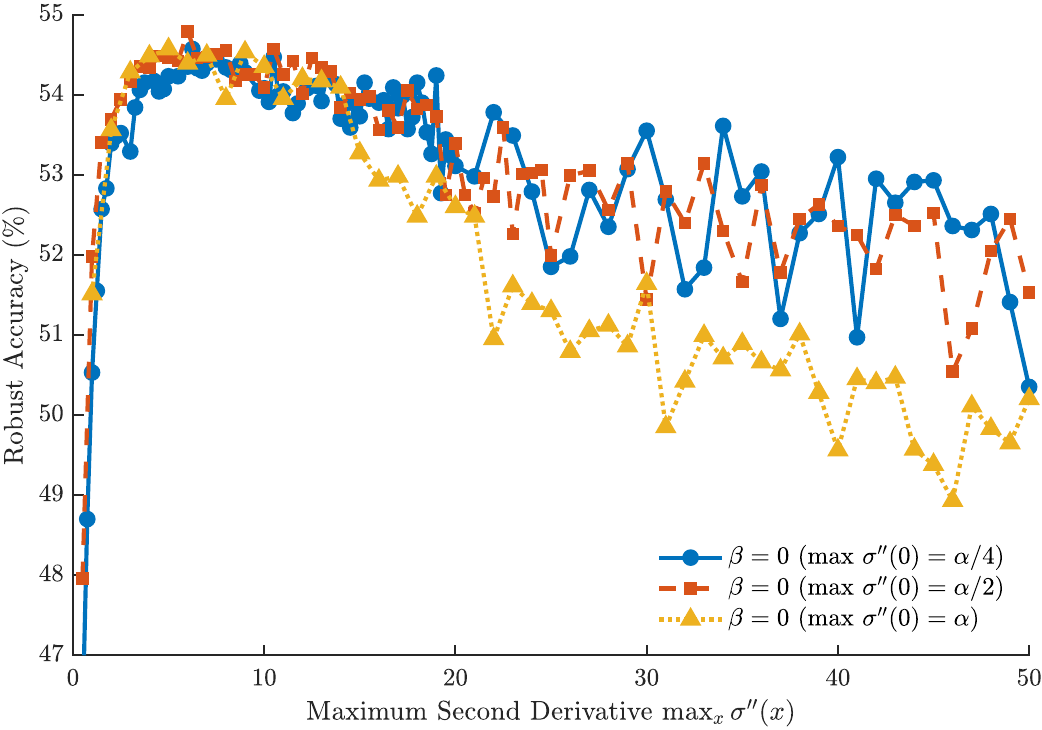}
\caption{Robustness vs. maximum curvature}
\label{fig:robustness_curvature}
\end{subfigure}
\caption{Robustness analysis of \yuaf\  activations  under DAJAT adversarial training on CIFAR-10 with ResNet-18.
(a) Robust accuracy vs. $\alpha$ for $\beta=0,1,2$, with $\alpha$ varying from 1 to 50 in steps of 1. The curves differ significantly across $\beta$ values, indicating that $\alpha$ alone does not determine robustness.
(b) Robust accuracy vs. $\max_x\sigma''(x)$ for the same models, where $\max_x\sigma''(x) = \alpha/4$ for $\beta=0$, $\alpha/2$ for $\beta=1$, and $\alpha$ for $\beta=2$. When $\max_x\sigma''(x) < 15$, the curves nearly overlap, showing that maximum curvature, not $\beta$, primarily governs robustness. 
}
\label{fig:robustness_analysis}
\end{figure*}

To empirically validate this theoretical relationship and study its impact on the overall loss landscape, we examine the Normalized Hessian diagonal $L_2$ norm, defined as:
\begin{equation}
\|\operatorname{diag}(\nabla^2_\theta \hat{L}(\theta))/p\|_2 = \sqrt{\frac{1}{p}\sum_{k=1}^p [\nabla^2_\theta \hat{L}(\theta)]_{kk}^2},
\label{eq:diag_norm_def}
\end{equation}
where $p$ is the total number of parameters. By summing squared values, this norm avoids the cancellation effects inherent in the trace and provides a robust measure of overall curvature intensity. It directly connects the activation second derivative—whose maximum value $\max|\sigma''|$ represents the peak curvature—to the sharpness of the loss landscape, enabling empirical investigation of how activation curvature shapes the loss geometry.

\section{Experiments}

\subsection{Experimental Setup}
We conduct experiments on CIFAR-10 and CIFAR-100 datasets \cite{krizhevsky2009learning} using two widely adopted network architectures: ResNet-18 \cite{he2016deep} and WideResNet-28-10 \cite{zagoruyko2016wide}, which provide a balanced trade-off between computational efficiency and representational capacity. To ensure the generality of our findings, we employ three distinct adversarial training methods: DAJAT \cite{addepalli2022efficient}, DKL \cite{cui2024decoupled}, and TRADES \cite{zhang2019theoretically}, all using $\ell_\infty$ perturbations with budget $\epsilon=8/255$ and following their original implementations. Model robustness is evaluated using \textbf{AutoAttack (AA)} \cite{croce2020reliable}, the community-standard ensemble attack, with all reported robust accuracy (RA) values obtained via AA at $\epsilon=8/255$. We systematically vary the $\max|\sigma''|$ parameter of \yuaf\  activations from 1 to 50 for $\beta \in \{0,1,2\}$, covering a wide range from insufficient to excessive curvature. Due to the substantial computational cost of adversarial training, we perform a single run of full adversarial training for each combination $(\alpha,\beta)$, noting that the results may exhibit some variability inherent to the training process. This comprehensive setup enables a thorough analysis of curvature effects while maintaining experimental feasibility.
\subsection{Core Finding: The Non-Monotonic Relationship Between Curvature and Robustness}


Figure~\ref{fig:robustness_analysis} presents our central experimental finding. Panel (a) shows robust accuracy versus $\alpha$ for $\beta=0,1,2$ under DAJAT training on CIFAR-10 with ResNet-18. The three curves exhibit distinct patterns; at the same $\alpha$ value, they yield substantially different robustness, suggesting that $\alpha$ alone cannot explain the robustness behavior across different functional forms. When transformed to the curvature coordinate system using $\max|\sigma''|$ values ($\alpha/4$, $\alpha/2$, and $\alpha$ for $\beta=0,1,2$ respectively), panel (b) reveals a unified phenomenon: all three curves nearly overlap when $\max|\sigma''| < 15$, demonstrating that maximum curvature—not the specific functional form—primarily determines adversarial robustness.

Three key observations emerge. First, for $\max|\sigma''| < 4$, increasing curvature enhances robustness across all $\beta$ values, as insufficient curvature limits nonlinear expressivity and rectification strength. Second, all \yuaf\  variants achieve peak robustness when $\max|\sigma''|$ falls between 4 and 10. Third, beyond $\max|\sigma''| > 10$, further curvature increases consistently degrade robustness, with $\beta=2$ (highest nonlinear complexity) showing the steepest decline, suggesting that activation functions with more complex curvature profiles (having $2\beta+1$ extrema in $\sigma''$) suffer more severely from excessive curvature. 
Together, these observations reveal a fundamental and quantifiable trade-off: activation functions must provide sufficient curvature ($\max|\sigma''| > 4$) for adequate nonlinear expressivity, yet must avoid excessive curvature ($\max|\sigma''| > 10$) to prevent the sharpening of the loss landscape that hinders robust generalization. The optimal range ($\max|\sigma''| \approx 4$–$10$) balances these competing demands.

\begin{figure}[h!]
\centering
\includegraphics[width=\linewidth]{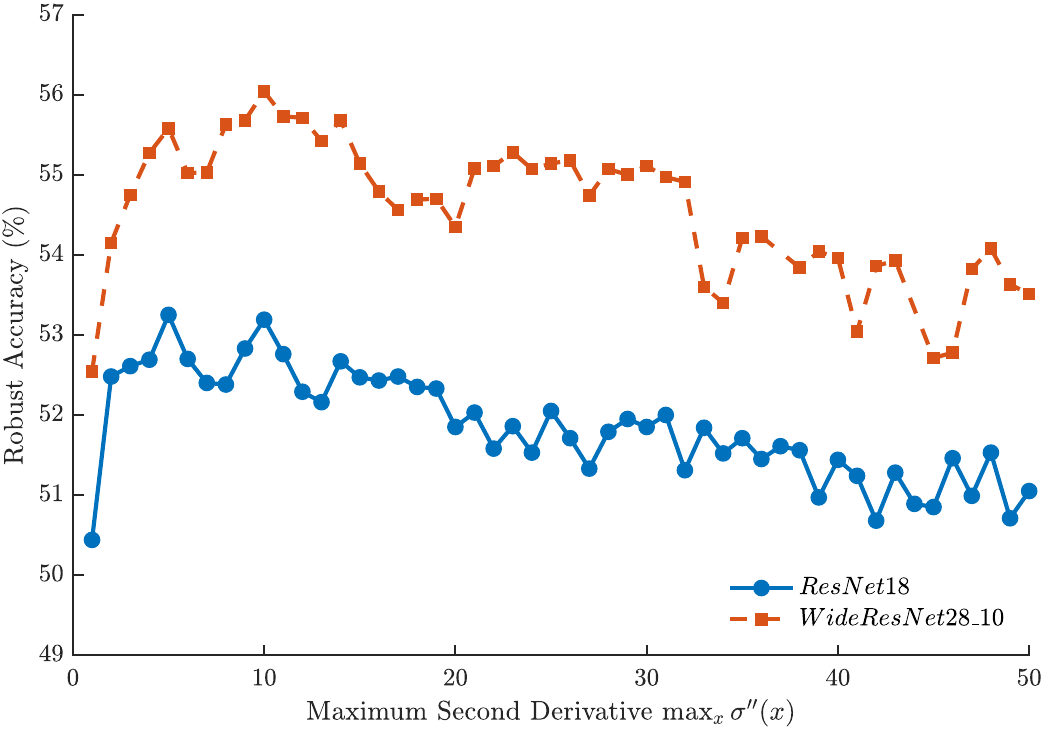}
\caption{Adversarial robustness vs. maximum second derivative $\max|\sigma''|$ for ResNet-18 and WideResNet-28-10 trained with TRADES on CIFAR-10 using \yuaf\  ($\beta=1$). Both architectures exhibit the same inverted-U relationship, with robustness peaking when $\max|\sigma''|$ lies between 4 and 10. The wider network (WideResNet-28-10) achieves higher absolute robust accuracy but follows an identical dependence on activation curvature, demonstrating that the identified optimal curvature range is architecture-agnostic.}
\label{fig:hessian_trace_vs_curvature_resnet_wideresnet}
\end{figure}

\subsection{Ablation Studies: Robustness Across Variations}

We conduct ablation studies to verify the consistency of our findings across different network architectures, adversarial training methods, and datasets.

\paragraph{Different Network Architectures} Figure~\ref{fig:hessian_trace_vs_curvature_resnet_wideresnet} compares ResNet-18 and WideResNet-28-10 under TRADES training on CIFAR-10. Both architectures exhibit the same inverted-U relationship between $\max|\sigma''|$ and robustness, with optimal performance occurring within $\max|\sigma''| \in [4,10]$. The wider network shows slightly higher absolute robustness but follows identical curvature dependence, confirming our findings are architecture-agnostic.

\begin{figure}[h!]
\centering
\includegraphics[width=\linewidth]{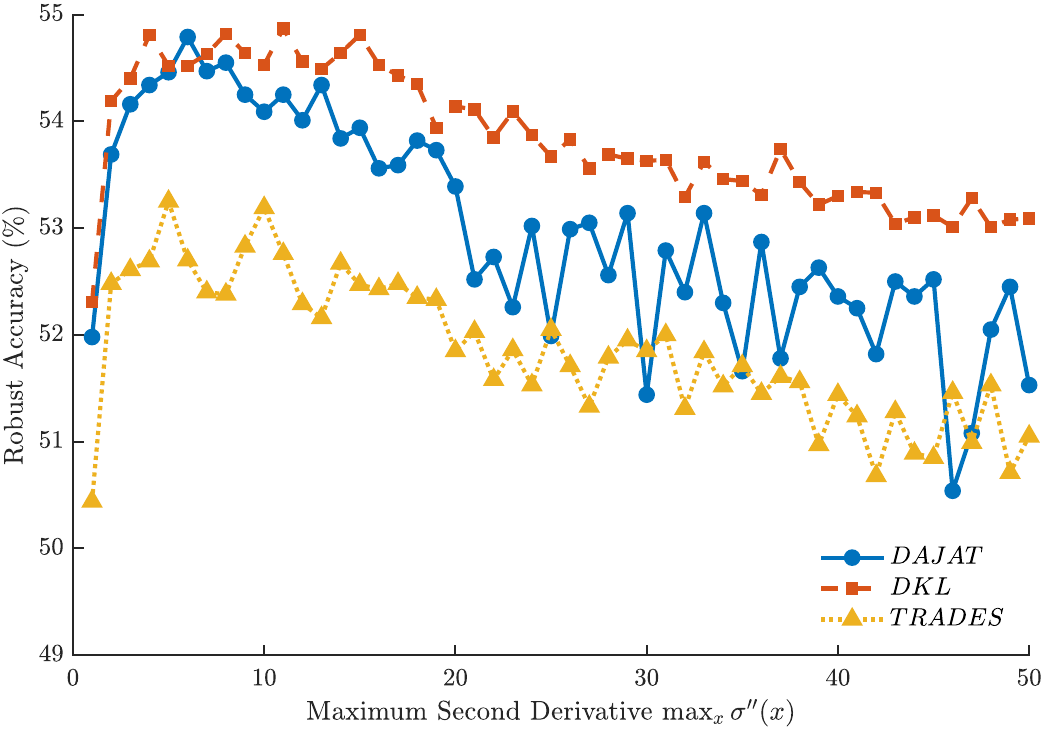}
\caption{Comparison of adversarial robustness vs. $\max|\sigma''|$ across three adversarial training methods—DAJAT, DKL, and TRADES—on CIFAR-10 with ResNet-18 and \yuaf\  ($\beta=1$). All methods exhibit the same qualitative behavior: robustness improves with curvature up to an optimum ($\max|\sigma''| \approx 4\text{--}10$) and declines thereafter.}
\label{fig:hessian_trace_vs_curvature_different_method}
\end{figure}

\paragraph{Different Adversarial Training Methods} Figure~\ref{fig:hessian_trace_vs_curvature_different_method} compares DAJAT, DKL, and TRADES on CIFAR-10 with ResNet-18 using \yuaf\  ($\beta=1$). All three methods exhibit identical qualitative behavior: robustness improves with $\max|\sigma''|$ up to 4-10, then declines. While absolute performance varies, the optimal curvature range remains consistent, demonstrating that our findings generalize across adversarial training paradigms.

\begin{figure}[h!]
\centering
\includegraphics[width=\linewidth]{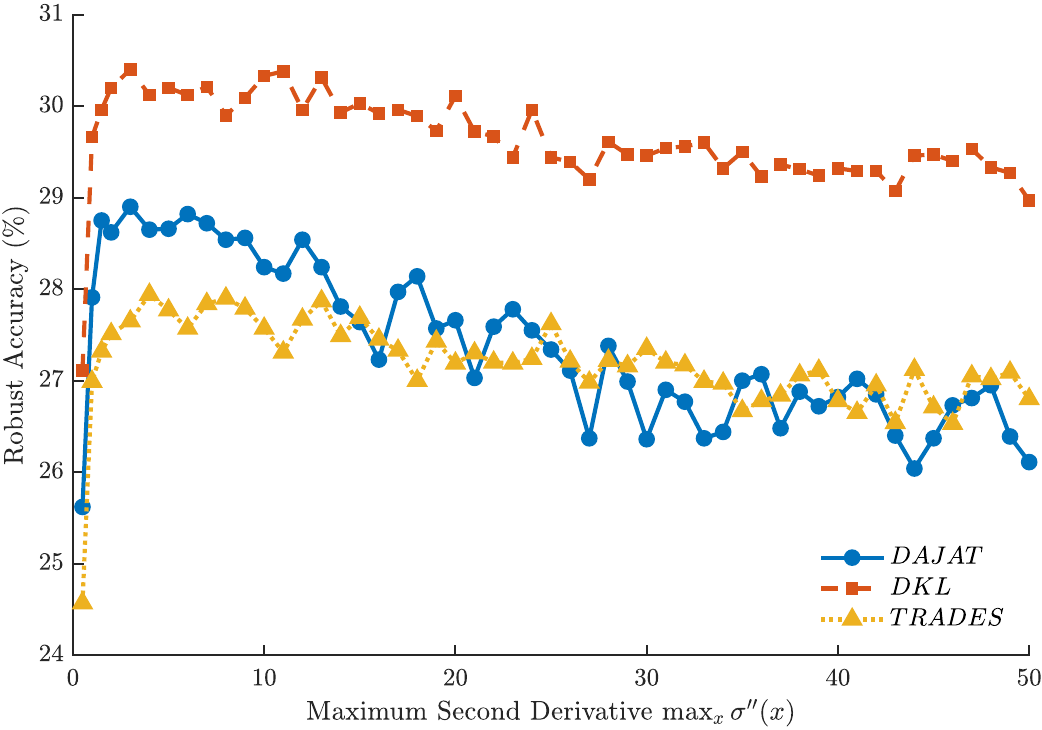}
\caption{Adversarial robustness vs. maximum second derivative $\max|\sigma''|$ evaluated on CIFAR-100 using ResNet-18 trained with DAJAT and \yuaf\  ($\beta=1$). Despite the increased complexity of CIFAR-100 (100 classes), the non-monotonic relationship persists, and optimal performance consistently occurs within $\max|\sigma''| \in [4, 10]$. }
\label{fig:hessian_trace_vs_curvature_different_dataset}
\end{figure}

\paragraph{Different Datasets} Figure~\ref{fig:hessian_trace_vs_curvature_different_dataset} extends evaluation to CIFAR-100 using ResNet-18 with DAJAT, DKL, and TRADES. Despite CIFAR-100's increased complexity (100 classes vs. 10), the relationship remains: robustness peaks at $\max|\sigma''| \in [4,10]$ and declines with further curvature increases. The optimal range shows remarkable consistency across datasets, though absolute robustness values are lower on the more challenging CIFAR-100.

Across architectures, training methods, and datasets, we consistently observe that optimal adversarial robustness occurs when $\max|\sigma''|$ falls within 4 to 10. This robustness suggests that the trade-off between rectification strength and curvature-induced sharpness represents a fundamental property of adversarial training, independent of implementation details.
\begin{figure*}[h!]
\centering
\begin{subfigure}{0.49\textwidth}
\centering
\includegraphics[width=\linewidth]{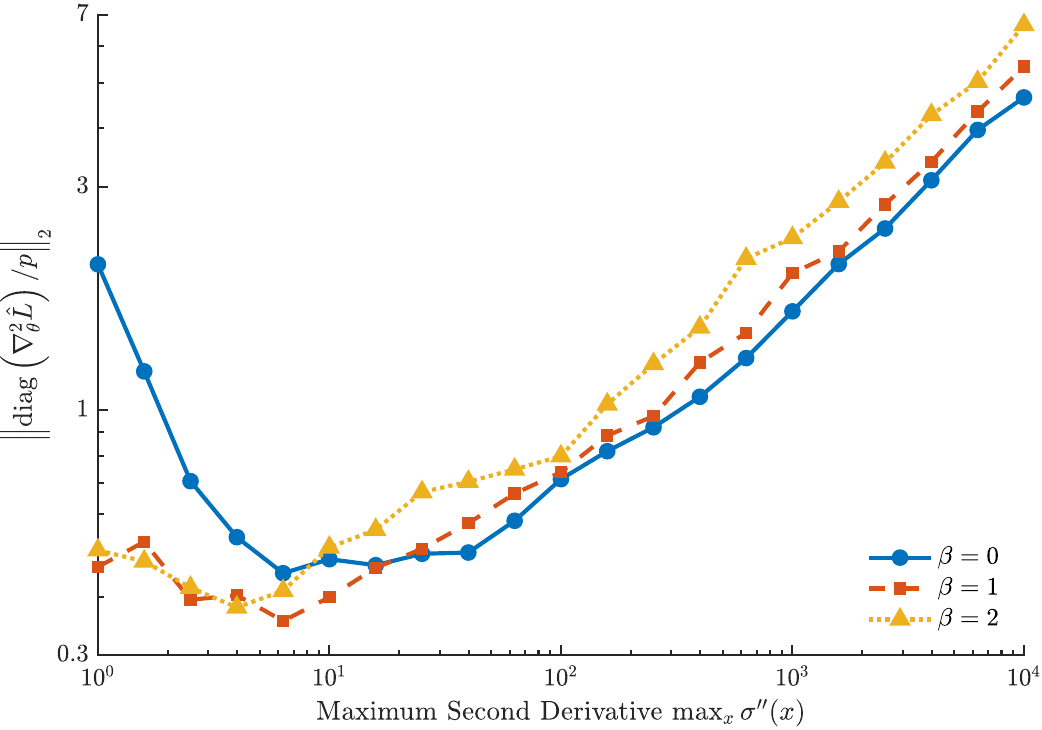}
\caption{Normalized Hessian Diagonal Norm vs. maximum curvature}
\label{fig:frobenius_norm}
\end{subfigure}
\hfill
\begin{subfigure}{0.49\textwidth}
\centering
\includegraphics[width=\linewidth]{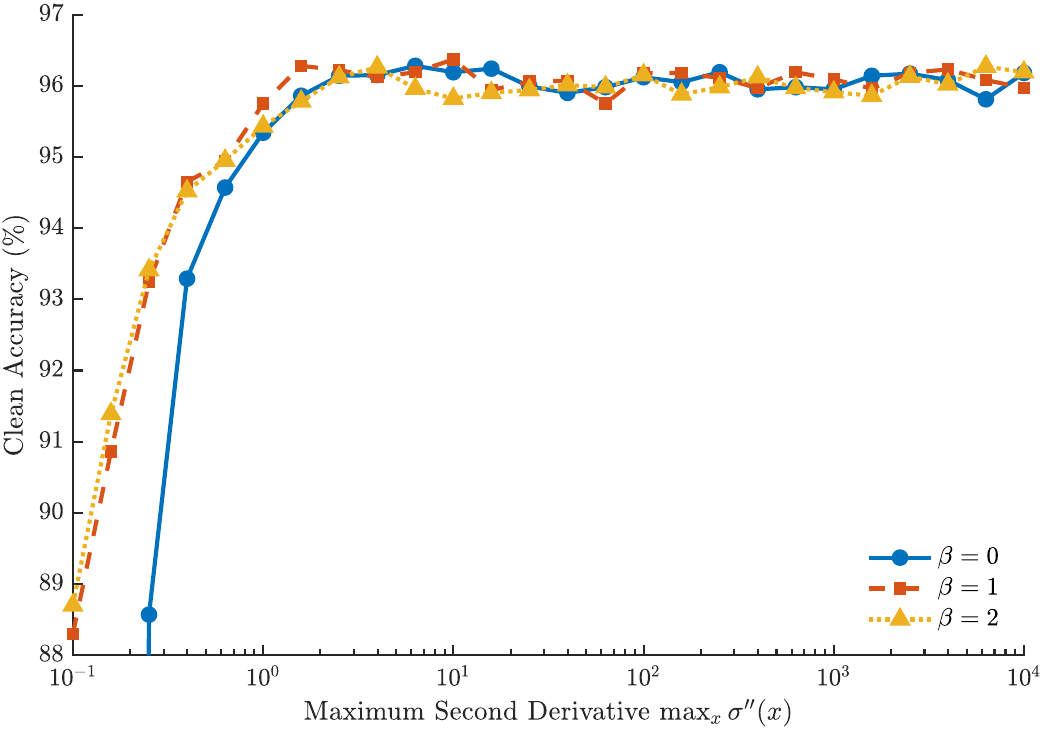}
\caption{Clean accuracy vs. maximum curvature}
\label{fig:frobenius_norm_clean_acc}
\end{subfigure}
\caption{(a) Measured normalized Hessian diagonal norm $\|\operatorname{diag}(\nabla^2_\theta \hat{L})/p\|_2$ versus $\max|\sigma''|$ for models trained with standard (non-adversarial) training  using the SGD optimizer on CIFAR-10 using ResNet-18 ($\beta=0,1,2$). Here, $p$ denotes the total number of model parameters. The plot reveals that when $\max|\sigma''|$ lies between 4 and 10, $\|\operatorname{diag}(\nabla^2_\theta \hat{L})/p\|_2$ reaches its minimum values across all $\beta$ values. When $\max|\sigma''|$ exceeds 10, further increases lead to marked rises in the norm, indicating sharper loss landscapes. Conversely, when $\max|\sigma''|$ falls below 4, further reductions in curvature also lead to increased norm, suggesting that insufficient curvature can similarly sharpen the landscape.
(b) Clean accuracy versus $\max|\sigma''|$ for the same set of models shown in (a) (standard training, ResNet-18 on CIFAR-10). When $\max|\sigma''|$ falls below 4, further reductions in curvature lead to a clear decrease in clean accuracy, demonstrating that insufficient curvature limits model capacity due to weak rectification and near-linear behavior. Together, these panels—measured from the same models under identical conditions—illustrate the dual effects of activation curvature: insufficient curvature impairs expressivity, while excessive curvature sharpens the loss landscape.}
\label{fig:hessian_diag_norm}
\end{figure*}


\begin{table}[h!]
\caption{Comparison of activation functions in adversarial training on CIFAR-10 with ResNet-18. All models are trained with DAJAT and evaluated using AutoAttack. \yuaf\  uses optimal parameters: $\max|\sigma''|=7$. Note: Leaky ReLU uses a negative slope of 0.01.}
\label{tab:activation_comparison}
\centering
\begin{tabular}{lccc}
\toprule
\makecell{Activation \\ Function} & \makecell{Standard \\ Accuracy} & \makecell{Robust \\ Accuracy } & $\max|\sigma''|$ \\
\midrule
ReLU & 85.23 & 51.37 & $\infty$  \\
Leaky ReLU & 84.52 & 51.24 & $\infty$  \\
ELU & 78.39 & 43.11 & 1.0 \\
GELU & 84.88 & 51.69 & 0.798 \\
Swish & 82.53 & 48.08 & 0.5 \\
Mish & 82.63 & 48.75 & 0.362 \\
\hline
\yuaf\ ($\beta=0$) & \textbf{86.65} & \textbf{54.57} & 7.0 \\
\yuaf\ ($\beta=1$) & \textbf{87.24} & \textbf{54.79} & 7.0 \\
\yuaf\ ($\beta=2$) & \textbf{86.08} & \textbf{54.49} & 7.0 \\
\bottomrule
\end{tabular}
\end{table}

\subsection{Comparison with State-of-the-Art Activation Functions}

Table~\ref{tab:activation_comparison} compares optimally tuned \yuaf\  uses optimal parameters: $\max|\sigma''|=7$ against popular activation functions under identical adversarial training settings. \yuaf\ ($\beta=0$) achieves 54.57\% robust accuracy, \yuaf\ ($\beta=1$) achieves 54.79\% robust accuracy, and \yuaf\ ($\beta=2$) achieves 54.49\% robust accuracy, outperforming all baselines. Notably, the best baseline GELU achieves 51.69\% robust accuracy, so the improvements are 2.88\%, 3.1\%, and 2.8\% for \yuaf\  with $\beta=0,1,2$ respectively.

Several important insights emerge. ReLU and LeakyReLU perform worse than \yuaf\ , consistent with their effectively infinite curvature at the kink creating sharp loss landscapes. Smooth activations like GELU, Swish, ELU, and Mish—with $\max|\sigma''| \leq 1.0$—also fall below \yuaf\ 's performance, suggesting their curvature is insufficient for optimal adversarial robustness. \yuaf\ 's $\max|\sigma''|=7$ falls within our identified optimal range (4-10), confirming that carefully controlled curvature balances expressivity and landscape flatness.

\subsection{Mechanism Analysis: Normalized Hessian Diagonal Norm and Loss Landscape Sharpness}
\label{sec:results:mechanism}

Figure~\ref{fig:hessian_diag_norm} presents our mechanism validation results, measuring the normalized Hessian Diagonal Norm $\|\operatorname{diag}(\nabla^2_\theta \hat{L})/p\|_2$, where $p$ denotes the total number of model parameters. The left panel shows the relationship between $\max|\sigma''|$ and this normalized metric, revealing a distinct non-monotonic pattern across all $\beta$ values.

A critical observation is that $\|\operatorname{diag}(\nabla^2_\theta \hat{L})/p\|_2$ reaches its minimum values when $\max|\sigma''|$ lies between 4 and 10. This range corresponds precisely to the region where we observe optimal adversarial robustness, suggesting that models with minimal normalized Hessian diagonal norm achieve the flattest loss landscapes and best robust generalization.

When $\max|\sigma''|$ exceeds 10, further increases lead to marked rises in the normalized Hessian diagonal norm, indicating that excessive activation curvature amplifies the overall curvature of the loss landscape along individual parameter directions. Conversely, when $\max|\sigma''|$ falls below 4, further reductions in curvature also lead to increased $\|\operatorname{diag}(\nabla^2_\theta \hat{L})/p\|_2$ across all $\beta$ values. This dual trend creates a clear U-shaped relationship: both insufficient and excessive curvature increase the normalized Hessian diagonal norm, while moderate curvature (4-10) minimizes it.

The right panel of Figure~\ref{fig:hessian_diag_norm} shows the relationship between $\max|\sigma''|$ and clean accuracy on standard training (without adversarial perturbations) using ResNet-18 on CIFAR-10. This analysis provides important insights into the model's fundamental capacity. We observe that when $\max|\sigma''|$ is below approximately 4, further reductions in curvature lead to a clear decrease in clean accuracy. This phenomenon provides direct evidence that when $\max|\sigma''|$ is too small, weak rectification leads to near-linear behavior, limiting the model's capacity to learn complex representations and reducing its overall expressivity.

These empirical measurements provide a comprehensive mechanistic explanation for the observed non-monotonic relationship between $\max|\sigma''|$ and adversarial robustness. The U-shaped pattern in the normalized Hessian diagonal norm reveals that both extremes of activation curvature sharpen the loss landscape: insufficient curvature ($\max|\sigma''| < 4$) leads to impaired expressivity and increased landscape sharpness, while excessive curvature ($\max|\sigma''| > 10$) directly amplifies the Hessian diagonal norm. The optimal $\max|\sigma''|$ range (4-10) achieves the crucial balance: it provides sufficient nonlinearity and rectification strength for effective feature learning (as evidenced by clean accuracy), while simultaneously minimizing the normalized Hessian diagonal norm to maintain a flat loss landscape conducive to robust generalization.

This trade-off explains why neither traditional smooth activations (with $\max|\sigma''| \leq 1.0$) nor ReLU (with effectively infinite curvature) achieve optimal robustness, and why carefully tuned activation functions with $\max|\sigma''|$ in the optimal range outperform both extremes. The normalized Hessian diagonal norm serves as a unifying metric that directly links activation curvature to loss landscape geometry, providing a mechanistic bridge between architectural choices and adversarial robustness.

\section{Conclusion}
\label{sec:conclusion}

This work systematically uncovers a fundamental relationship between activation function curvature and adversarial robustness, revealing a non-monotonic pattern where both insufficient and excessive curvature degrade robust performance, with an optimal range observed when the maximum second derivative $\max|\sigma''|$ falls between 4 and 10. Theoretically, we establish that the activation second derivative $\sigma''$ directly influences individual diagonal elements of the loss Hessian $[\nabla^2_\theta \hat{L}(\theta)]_{kk}$, providing a mathematical foundation for understanding how activation functions shape loss landscape curvature. Empirically, we observe that when $\max|\sigma''|$ exceeds 10, the Hessian diagonal norm $\|\operatorname{diag}(\nabla^2_\theta \hat{L}(\theta))\|_2$ exhibits a clear upward trend, indicating a sharper loss landscape that hinders robust generalization—this experimentally observed correlation offers a mechanistic explanation for the degradation in adversarial robustness beyond the optimal curvature range. These findings hold consistently across diverse network architectures, datasets, and adversarial training methods, establishing curvature control as a fundamental principle for designing robust neural networks and providing both theoretical insights and practical guidance for activation function selection in adversarial settings.


\clearpage

{
    \small
    \bibliographystyle{ieeenat_fullname}
    \bibliography{main}
}

\clearpage

\appendix
\section{Detailed Derivation: Relationship Between Hessian Diagonal Elements and Activation Second Derivatives}
\label{app:hessian_diag_derivation}

\subsection{Notation and Variable Definitions}
\label{app:notation}

For clarity, we define all symbols and variables used in this derivation:

\begin{itemize}
    \item $x \in \mathbb{R}^d$: Input feature vector.
    \item $y \in \mathbb{R}$: Target value (scalar regression).
    \item $L$: Total number of layers (including output layer).
    \item $n_l$: Number of neurons in layer $l$, with $n_0 = d$ (input dimension) and $n_L = 1$ (output).
    \item $W^{(l)} \in \mathbb{R}^{n_l \times n_{l-1}}$: Weight matrix for layer $l$.
    \item $b^{(l)} \in \mathbb{R}^{n_l}$: Bias vector for layer $l$.
    \item $z^{(l)} \in \mathbb{R}^{n_l}$: Pre-activation vector at layer $l$, with $z^{(l)} = W^{(l)} h^{(l-1)} + b^{(l)}$.
    \item $h^{(l)} \in \mathbb{R}^{n_l}$: Post-activation vector at layer $l$, with $h^{(l)} = \sigma(z^{(l)})$ and $h^{(0)} = x$.
    \item $\sigma(\cdot)$: Activation function, applied elementwise and assumed twice differentiable.
    \item $\sigma'(\cdot)$, $\sigma''(\cdot)$: First and second derivatives of $\sigma$.
    \item $f \in \mathbb{R}$: Network output, $f = z^{(L)} = W^{(L)} h^{(L-1)} + b^{(L)}$ (linear output layer).
    \item $\hat{L}(\theta) = \frac{1}{2}(f - y)^2$: Squared loss function.
    \item $\theta_k$: Any individual network parameter (weight or bias).
    \item $\delta^{(l)}_i = \frac{\partial f}{\partial z^{(l)}_i}$: Backpropagated gradient for neuron $i$ in layer $l$.
    \item $c_k = \frac{\partial z^{(l)}_i}{\partial \theta_k}$: Derivative of pre-activation w.r.t. parameter $\theta_k$ associated with neuron $i$ in layer $l$.
    \item $D^{(l)}_i = \frac{\partial \delta^{(l)}_i}{\partial z^{(l)}_i}$: Derivative of backpropagated gradient w.r.t. its own pre-activation.
    \item $[\nabla^2_\theta \hat{L}]_{kk}$: Diagonal element of the loss Hessian corresponding to parameter $\theta_k$.
\end{itemize}

\subsection{General Framework: Hessian Diagonal Element Decomposition}
\label{app:general_framework}

For any parameter $\theta_k$ and squared loss $\hat{L}(\theta) = \frac{1}{2}(f-y)^2$, the Hessian diagonal element decomposes via the chain rule as:
\begin{align}
    [\nabla^2_\theta \hat{L}]_{kk} 
    &= \frac{\partial^2 \hat{L}}{\partial \theta_k^2} 
    = \frac{\partial}{\partial \theta_k} \left( (f-y) \frac{\partial f}{\partial \theta_k} \right) \nonumber \\
    &= \left( \frac{\partial f}{\partial \theta_k} \right)^2 + (f-y) \frac{\partial^2 f}{\partial \theta_k^2}.
    \label{eq:diag_decomp_general}
\end{align}

Thus, understanding $[\nabla^2_\theta \hat{L}]_{kk}$ reduces to analyzing $\frac{\partial f}{\partial \theta_k}$ and $\frac{\partial^2 f}{\partial \theta_k^2}$, where the latter depends on $\sigma''$.

\subsection{Single Hidden Layer Network}
\label{app:single_hidden_layer}

We first analyze a network with one hidden layer ($L=2$) to build intuition.

\subsubsection{Network Architecture}
The network has input $x$, hidden layer with $m$ neurons, and linear output:
\begin{align}
    & z_i = \sum_{j=1}^d W^{(1)}_{ij} x_j + b^{(1)}_i, \quad i=1,\dots,m, \nonumber \\
    & h_i = \sigma(z_i), \nonumber \\
    & f = \sum_{i=1}^m W^{(2)}_i h_i + b^{(2)}.
    \label{eq:single_layer_forward}
\end{align}

\subsubsection{Output Layer Parameters}
For output layer parameters, $f$ is linear, so second derivatives vanish:
\begin{align}
    \frac{\partial^2 f}{\partial (W^{(2)}_i)^2} = 0, \quad
    \frac{\partial^2 f}{\partial (b^{(2)})^2} = 0.
\end{align}
Thus, from \eqref{eq:diag_decomp_general}:
\begin{align}
    [\nabla^2_\theta \hat{L}]_{W^{(2)}_i,W^{(2)}_i} &= h_i^2, \quad
    [\nabla^2_\theta \hat{L}]_{b^{(2)},b^{(2)}} = 1.
\end{align}
These elements depend only on activations, not on $\sigma''$.

\subsubsection{Hidden Layer Weights $W^{(1)}_{ij}$}
For hidden weights, we compute:
\begin{align}
    &\frac{\partial f}{\partial W^{(1)}_{ij}} 
    = \frac{\partial f}{\partial h_i} \frac{\partial h_i}{\partial z_i} \frac{\partial z_i}{\partial W^{(1)}_{ij}} 
    = W^{(2)}_i \sigma'(z_i) x_j, \\
    &\frac{\partial^2 f}{\partial (W^{(1)}_{ij})^2} 
    = \frac{\partial}{\partial W^{(1)}_{ij}} \left( W^{(2)}_i \sigma'(z_i) x_j \right) 
    = W^{(2)}_i \sigma''(z_i) x_j^2.
\end{align}
Substituting into \eqref{eq:diag_decomp_general}:
\begin{align}
    & [\nabla^2_\theta \hat{L}]_{W^{(1)}_{ij},W^{(1)}_{ij}} = \nonumber  \\
     & \left( W^{(2)}_i \sigma'(z_i) x_j \right)^2 +  (f-y) W^{(2)}_i \sigma''(z_i) x_j^2.
    \label{eq:diag_weight_single}
\end{align}

\subsubsection{Hidden Layer Biases $b^{(1)}_i$}
Similarly:
\begin{align}
    &\frac{\partial f}{\partial b^{(1)}_i} 
    = W^{(2)}_i \sigma'(z_i), \\
    &\frac{\partial^2 f}{\partial (b^{(1)}_i)^2} 
    = W^{(2)}_i \sigma''(z_i).
\end{align}
Thus:
\begin{align}
    & [\nabla^2_\theta \hat{L}]_{b^{(1)}_i,b^{(1)}_i} 
    = \nonumber \\
    & \left( W^{(2)}_i \sigma'(z_i) \right)^2 +   (f-y) W^{(2)}_i \sigma''(z_i).
    \label{eq:diag_bias_single}
\end{align}


\subsection{Extension to Deep Networks}
\label{app:deep_networks}

We now generalize to $L$-layer fully connected networks.

\subsubsection{Forward and Backward Propagation}
For layer $l = 1,\dots,L-1$:
\begin{align}
    & z^{(l)} = W^{(l)} h^{(l-1)} + b^{(l)}, \nonumber \\
    & h^{(l)} = \sigma(z^{(l)}), \quad 
    h^{(0)} = x, \nonumber \\
    & f = z^{(L)} = W^{(L)} h^{(L-1)} + b^{(L)}.
\end{align}
The backpropagated gradients are:
\begin{align}
    \delta^{(L)} &= 1, \quad 
    \delta^{(l)}_i = \sigma'(z^{(l)}_i) \sum_{t=1}^{n_{l+1}} \delta^{(l+1)}_t W^{(l+1)}_{ti}.
    \label{eq:backprop_deep}
\end{align}

\subsubsection{Diagonal Element for General Parameter $\theta_k$}
Let $\theta_k$ be associated with neuron $i$ in layer $l$. Define:
\begin{align}
    c_k = \frac{\partial z^{(l)}_i}{\partial \theta_k} = 
    \begin{cases}
        h^{(l-1)}_j & \text{if } \theta_k = W^{(l)}_{ij}, \\
        1 & \text{if } \theta_k = b^{(l)}_i.
    \end{cases}
\end{align}
Then the first derivative is:
\begin{align}
    \frac{\partial f}{\partial \theta_k} = \delta^{(l)}_i c_k.
    \label{eq:first_deriv_general}
\end{align}
For the second derivative, since $c_k$ is independent of $\theta_k$ (for weights, $h^{(l-1)}_j$ does not depend on $W^{(l)}_{ij}$; for biases, $c_k=1$ is constant), we have:
\begin{align}
    \frac{\partial^2 f}{\partial \theta_k^2} 
    &= \frac{\partial}{\partial \theta_k} (\delta^{(l)}_i c_k) 
    = \left( \frac{\partial \delta^{(l)}_i}{\partial \theta_k} \right) c_k \nonumber \\
    &= \left( \frac{\partial \delta^{(l)}_i}{\partial z^{(l)}_i} \frac{\partial z^{(l)}_i}{\partial \theta_k} \right) c_k 
    = \frac{\partial \delta^{(l)}_i}{\partial z^{(l)}_i} c_k^2.
    \label{eq:second_deriv_general}
\end{align}

\subsubsection{Computing $D^{(l)}_i = \frac{\partial \delta^{(l)}_i}{\partial z^{(l)}_i}$}
From \eqref{eq:backprop_deep}, we derive a recurrence for $D^{(l)}_i$:
\begin{align}
    D^{(l)}_i 
    = & \frac{\partial}{\partial z^{(l)}_i} \left[ 
        \sigma'(z^{(l)}_i) \sum_{t=1}^{n_{l+1}} \delta^{(l+1)}_t W^{(l+1)}_{ti} 
    \right] \nonumber \\
    = & \sigma''(z^{(l)}_i) \sum_{t=1}^{n_{l+1}} \delta^{(l+1)}_t W^{(l+1)}_{ti} + \nonumber \\
    &  \sigma'(z^{(l)}_i) \sum_{t=1}^{n_{l+1}} \frac{\partial \delta^{(l+1)}_t}{\partial z^{(l)}_i} W^{(l+1)}_{ti}.
    \label{eq:D_deriv_initial}
\end{align}

To compute $\frac{\partial \delta^{(l+1)}_t}{\partial z^{(l)}_i}$, note that $\delta^{(l+1)}_t$ depends on $z^{(l)}_i$ via $z^{(l+1)}_t$:
\begin{align}
    & z^{(l+1)}_t = \sum_{j=1}^{n_l} W^{(l+1)}_{tj} \sigma(z^{(l)}_j) + b^{(l+1)}_t, \\
    & \frac{\partial z^{(l+1)}_t}{\partial z^{(l)}_i} = W^{(l+1)}_{ti} \sigma'(z^{(l)}_i).
\end{align}
Thus, by the chain rule:
\begin{align}
    \frac{\partial \delta^{(l+1)}_t}{\partial z^{(l)}_i} 
    &= \frac{\partial \delta^{(l+1)}_t}{\partial z^{(l+1)}_t} \frac{\partial z^{(l+1)}_t}{\partial z^{(l)}_i} 
    = D^{(l+1)}_t \cdot W^{(l+1)}_{ti} \sigma'(z^{(l)}_i).
    \label{eq:delta_cross_deriv}
\end{align}

Substituting \eqref{eq:delta_cross_deriv} into \eqref{eq:D_deriv_initial}:
\begin{align}
    D^{(l)}_i &= \sigma''(z^{(l)}_i) S^{(l)}_i 
    + [\sigma'(z^{(l)}_i)]^2 \sum_{t=1}^{n_{l+1}} D^{(l+1)}_t (W^{(l+1)}_{ti})^2,
    \label{eq:D_recurrence}
\end{align}
where $S^{(l)}_i = \sum_{t=1}^{n_{l+1}} \delta^{(l+1)}_t W^{(l+1)}_{ti}$.
From \eqref{eq:backprop_deep}, when $\sigma'(z^{(l)}_i) \neq 0$,
\begin{align}
    S^{(l)}_i = \frac{\delta^{(l)}_i}{\sigma'(z^{(l)}_i)}.
\end{align}
Thus, the recurrence becomes:
\begin{align}
    D^{(l)}_i & = \sigma''(z^{(l)}_i) \frac{\delta^{(l)}_i}{\sigma'(z^{(l)}_i)} \nonumber \\
    & + [\sigma'(z^{(l)}_i)]^2 \sum_{t=1}^{n_{l+1}} D^{(l+1)}_t (W^{(l+1)}_{ti})^2.
    \label{eq:D_recurrence_final}
\end{align}
The base case is $D^{(L)} = 0$ because $\delta^{(L)} = 1$ is constant.

\subsubsection{Recursive Expansion of $D^{(l)}_i$}
Expanding \eqref{eq:D_recurrence_final} recursively reveals that $D^{(l)}_i$ is a weighted sum of $\sigma''$ terms from all subsequent layers:

\begin{equation}
\begin{split}
D^{(l)}_i = \sum_{r=l}^{L-1} \sum_{\substack{\text{paths } P: \\ (l,i) \to (r,j)}} 
& \sigma''(z^{(r)}_j) \cdot \frac{\delta^{(r)}_j}{\sigma'(z^{(r)}_j)} \\
& \times \prod_{s=l}^{r-1} [\sigma'(z^{(s)}_{i_s})]^2 (W^{(s+1)}_{i_{s+1}, i_s})^2.
\end{split}
\label{eq:D_expansion}
\end{equation}
where each path $P$ connects neuron $(l,i)$ to neuron $(r,j)$ via intermediate neurons. The product term represents attenuation through weights and activation derivatives.

\subsubsection{Final Expression for Deep Networks}
Combining \eqref{eq:diag_decomp_general}, \eqref{eq:first_deriv_general}, \eqref{eq:second_deriv_general}, and \eqref{eq:D_recurrence_final}, we obtain the general Hessian diagonal element:

\begin{align}
    [\nabla^2_\theta \hat{L}]_{kk} 
    &= (\delta^{(l)}_i c_k)^2 \nonumber \\
    &\quad + (f-y) c_k^2 \Biggl[ 
        \sigma''(z^{(l)}_i) \frac{\delta^{(l)}_i}{\sigma'(z^{(l)}_i)} \nonumber \\
    &\qquad + [\sigma'(z^{(l)}_i)]^2 \sum_{t=1}^{n_{l+1}} D^{(l+1)}_t (W^{(l+1)}_{ti})^2 \Biggr].
    \label{eq:diag_general_deep}
\end{align}
Alternatively, using the expanded form \eqref{eq:D_expansion}:

\begin{multline}
    [\nabla^2_\theta \hat{L}]_{kk} 
    = (\delta^{(l)}_i c_k)^2 + (f-y) c_k^2 \\
    \times \sum_{r=l}^{L-1} \sum_{\substack{\text{paths } P: \\ (l,i) \to (r,j)}} 
        \sigma''(z^{(r)}_j) \cdot \frac{\delta^{(r)}_j}{\sigma'(z^{(r)}_j)} \\
    \times \prod_{s=l}^{r-1} [\sigma'(z^{(s)}_{i_s})]^2 (W^{(s+1)}_{i_{s+1}, i_s})^2.
    \label{eq:diag_expanded}
\end{multline}

\subsection{Key Theoretical Insights}
\label{app:insights}

From our derivations, we highlight the following points:

\begin{itemize}
    \item \textbf{Linear Dependence on $\sigma''$:} The Hessian diagonal element depends linearly on $\sigma''$ at multiple layers, as seen in \eqref{eq:diag_expanded}.
    
    \item \textbf{Amplification by Prediction Error:} The $\sigma''$ contributions are scaled by $(f-y)$, meaning larger prediction errors amplify the effect of activation second derivatives on the Hessian diagonal.
    
    
\end{itemize}

\subsection{Computational Overhead Analysis}
\label{app:computational_overhead}

We analyze the computational overhead of \yuaf\  compared to standard activation functions under identical experimental settings: ResNet-18 on CIFAR-10 with batch size 128 for 20 epochs of standard training, measured on an NVIDIA 2080Ti GPU. Table~\ref{tab:computational_overhead} summarizes GPU memory usage and training time for various activation functions.

\begin{table}[h]
\centering
\begin{tabular}{lcc}
\hline
\makecell{Activation\\ Function} & \makecell{GPU Memory \\(GB)} & \makecell{Training Time \\(min)} \\ \hline
ReLU & 0.73 & 5.88 \\
LeakyReLU & 0.73 & 5.92 \\
ELU & 0.73 & 5.83 \\
SELU & 0.73 & 5.88 \\
GELU & 0.99 & 7.92 \\
Mish & 1.25 & 6.98 \\
Softplus & 0.99 & 5.87 \\
Swish & 0.99 & 5.88 \\
\yuaf\  ($\beta=0$) & 0.99 & 6.30 \\
\yuaf\  ($\beta=1$) & 1.25 & 6.93 \\
\yuaf\  ($\beta=2$) & 2.30 & 8.86 \\ \hline
\end{tabular}
\caption{Computational overhead comparison of activation functions under identical training settings (ResNet-18, CIFAR-10, batch size 128, 20 epochs). GPU memory usage and training time are measured on an NVIDIA 2080Ti.}
\label{tab:computational_overhead}
\end{table}

The results reveal several important patterns. First, \yuaf\  with $\beta=0$ exhibits memory usage comparable to GELU (0.99 GB) and training time between ReLU and GELU (6.30 min), representing a modest increase over simpler activations. Second, \yuaf\  with $\beta=1$ shows memory usage equivalent to Mish (1.25 GB) while maintaining training time between ReLU and GELU (6.93 min). Third, \yuaf\  with $\beta=2$ incurs significantly higher memory (2.30 GB) and training time (8.86 min), indicating that higher-order variants introduce substantial computational overhead.

Critically, our experiments demonstrate that adversarial robustness is primarily determined by $\max|\sigma''|$ rather than the specific $\beta$ value. Since $\beta=0$ and $\beta=1$ variants achieve similar robustness performance to $\beta=2$ when $\max|\sigma''|$ is properly tuned, there is no robustness advantage to using the more computationally expensive $\beta=2$ variant. Therefore, in practice, we recommend using \yuaf\  with $\beta=0$ or $\beta=1$, which provide optimal robustness without significant computational penalty compared to established activation functions like GELU and Mish.

\end{document}